\documentclass{article}
\usepackage{enumitem}
\usepackage{ijcai25}
\usepackage{amsthm}
\usepackage{times}
\usepackage{soul}
\usepackage{url}
\usepackage[hidelinks]{hyperref}
\usepackage[utf8]{inputenc}
\usepackage[small]{caption}
\usepackage{subfigure}

\usepackage{graphicx}
\usepackage{amsmath}
\usepackage{amsthm}
\usepackage{booktabs}
\usepackage{algorithm}
\usepackage{algorithmic}
\usepackage[switch]{lineno}
\usepackage{bm}
\usepackage{amsfonts}
\usepackage{amssymb}
\usepackage{array}

\usepackage{xcolor}
\usepackage{multirow}
\newtheorem{definition}{Definition}

\title{Resolving Latency and Inventory Risk in
Market Making with \\ Reinforcement Learning}
\title{Resolving Latency and Inventory Risk in
Market Making with \\ Reinforcement Learning}
\title{Resolving Latency and Inventory Risk in
Market Making with \\ Reinforcement Learning}

\author{
Junzhe Jiang$^{1}$\thanks{Equal contribution, alphabetical order}~, Chang Yang$^{1}$\footnotemark[1]~, \textbf{Xinrun Wang$^{2}$\footnotemark[2], Zhiming Li$^{3}$, Xiao Huang$^{1}$, Bo Li$^{1}$ }\\
$^1$The Hong Kong Polytechnic University, $^2$Singapore Management University,\\ $^3$Nanyang Technological University \\
\texttt{junzhe.jiang@connect.polyu.hk, xrwang@smu.edu.sg}}

\begin{document}

\maketitle

\begin{abstract}
The latency of the exchanges in Market Making (MM) is inevitable due to hardware limitations, system processing times, delays in receiving data from exchanges, the time required for order transmission to reach the market, etc.
Existing reinforcement learning (RL) methods for Market Making (MM) overlook the impact of these latency, which can lead to unintended order cancellations due to price discrepancies between decision and execution times and result in undesired inventory accumulation, exposing MM traders to increased market risk. Therefore, these methods cannot be applied in real MM scenarios. To address
these issues, we first build a realistic MM environment with random delays of 30-100 milliseconds for order placement and market information reception, and implement a batch matching mechanism that collects orders within every 500 milliseconds before matching them all at once, simulating the batch auction mechanisms adopted by some exchanges.
Then, we propose \textsc{Relaver}, an RL-based method for MM to tackle the latency and inventory risk issues. The three main contributions of \textsc{Relaver} are: i) we introduce an augmented state-action space that incorporates order hold time alongside price and volume, enabling \textsc{Relaver} to optimize execution strategies under latency constraints and time-priority matching mechanisms, ii) we leverage dynamic programming
(DP) to guide the exploration of RL training for better policies,
iii) we train a market trend predictor, which can guide the agent
to intelligently adjust the inventory to reduce the risk. Extensive experiments and ablation studies on four real-world datasets demonstrate that \textsc{Relaver} significantly improves the performance of state-of-the-art RL-based MM strategies across multiple metrics.
\end{abstract}

\section{Introduction}
Market making (MM) is the process where the firms and individuals continuously place buy and sell orders for financial assets such as stocks, bonds, or currencies~\cite{madhavan2000market}. For example, Citadel Securities, one of the largest U.S. market makers, reportedly handles handles around 25\% of all US equities trades~\cite{doherty2023citadel}. MM participates more than 60\% exchanges in the main cryptocurrency market~\cite{hk2022observations}. MM can provide liquidity and reduced the volatility to the financial markets. However, due to the complex market dynamics, it is extremely challenging for market makers (MMers) to make the optimal decisions to maximize the utilities while avoiding the various risks. Therefore, the latency of the exchanges prohibits the real deployment of existing methods into the real MM scenarios. 

\begin{figure}[ht]
    \centering
    \includegraphics[width=0.9\linewidth]{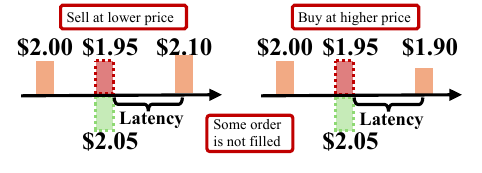}
    \vspace{-10pt}
    \caption{An example about the influence of latency}
    \label{fig:example}
\end{figure}
Recent methods leverage Reinforcement Learning (RL) for MM. Different from traditional methods based on the rules and mathematical models, RL-based MM can handle the complex market dynamics and adapt to the change of the market trends~\cite{niu2023imm}. However, previous methods usually do not consider \emph{the latency of the exchanges} in MM due to the hardware limitations, system processing time, delays in receiving data from exchanges, and the time required for order transmission to reach the market, etc. The overlook of the latency of the exchanges will lead to the failure of the orders, thus accumulate the inventory risks. For example, suppose that MMer A decides to buy apples at \$1.95 and sell them at \$2.05, given the current price at \$2. However, due to the latency, the apple price is changed to \$2.10 when the order reaches the system. Therefore, A sells the apples below the market prices, i.e., $\$2.05 < \$2.10$, and the buy order will not be filled as $\$1.95 <\$2.10$. Therefore, due to the latency, MMer A misses the chance to sell the apples with higher prices. When the apple prices significantly go down during the latency time, MMer A's sell order will inevitably be not filled, thus leading to the undesirable accumulation of the potential inventory risk, as depicted in Figure~\ref{fig:example}.

To address these issues, we first build a realistic MM environments, which introduce the random delays of the placing orders to model the latency of the exchanges and the batch matching mechanism to match the orders accumulated in a period, where the batch matching mechanism is used in the real MM scenarios. Then, we propose \textbf{\textsc{Relaver}}: \textbf{RE}solving \textbf{LA}ency and in\textbf{VE}ntory \textbf{R}isk, an RL-based method for MM to tackle the latency and inventory risk issues. The three main contributions of \textsc{Relaver} are: i) we introduce an augmented state-action space that incorporates order hold time alongside price and volume, enabling \textsc{Relaver} to optimize execution strategies under latency constraints and time-priority matching mechanisms, ii) we train the MM policy with RL, e.g., PPO, and leverage the teacher computed by dynamic programming
(DP) to guide the exploration of RL training for better policies,
iii) we train a market trend predictor, where given the predicted market trend, specific rules is applied to guide the agent
to intelligently adjust the inventory to reduce the risk. Extensive experiments and ablation studies on four real-world datasets demonstrate that \textsc{Relaver} significantly improves the performance of state-of-the-art RL-based MM strategies across multiple metrics.

\textbf{Related Work.} Traditional MM approaches \cite{ho1981optimal,labadie2013high,gueant2013dealing} treat MM as a stochastic optimal control problem using Poisson processes for order arrivals and price changes. \cite{guilbaud2013optimal} modeled asset midprice as Brownian motion and solved it using Hamilton-Jacobi-Bellman equations, while \cite{cartea2015risk} incorporated sophisticated risk metrics. However, these models rely on idealized processes rather than empirical data. 
Recent years have seen increased adoption of RL in MM \cite{kumar2020deep,ganesh2019reinforcement,chung2022market,gavsperov2021market,jerome2022market}. One dominant approach uses single-price level strategies \cite{sadighian2019deep,gueant2019deep,xu2022performance,beysolow2019market}, but these lack flexibility and fail to manage inventory risks effectively. Another approach employs multi-price level strategies \cite{chakraborty2011market,abernethy2013adaptive}, allowing simultaneous decisions on prices and volumes. \cite{niu2023imm} designed a low-dimensional representation for complex multi-price strategies, though frequent order cancellations remain a concern.
For realistic MM decisions, multi-price level strategies with fine-grained actions are needed to consider long-term market fluctuations and reserve orders for better margins. Additionally, previous work often overlooked market latency, leading to unrealistic assumptions \cite{gao2020optimal}. Please refer to Appendix~\ref{app:related_work} for a more thorough discussion.

\section{Problem Statement}

In this section, we introduce fundamental financial concepts utilized in simulating the market-making process and formulate a more realistic environment as a Markov Decision Process (MDP) framework for MM.

\subsection{Market Making}
        
In this section, we will introduce the preliminaries of Market Making (MM) including the definitions of latency, limited order book and order matching mechanism to facilitate the understanding. 

In market making (MM), \textbf{\emph{latency}} is defined as the time $s^{l}_{t}$ between the placement of an order and its execution. This latency is typically modeled as a random variable $\mathcal{U}(\cdot)$, i.e., $s^{l}_{t} \sim \mathcal{U}(\cdot)$. Consequently, the execution time of an order is given by $t_{e} = t + s^{l}_{t}$, where $t$ represents the time of order placement. An \textbf{\emph{order}} in MM is denoted as a tuple $o_t = (p_t, q_t, t_{e}, \tau)$, where $p_t$ and $q_t$ are the price and quantity of the order, $t_{e}$ is the execution time, and $\tau \in \{\text{ask}, \text{bid}\}$ specifies whether the order is an ask or bid.

The \textbf{\emph{Limit Order Book (LOB)}}, denoted as $OB_t$, is the collection of all active orders at a given time $t$. An $m$-level LOB at time $t$ is represented as $OB_t = \{o_1, o_2, o_3, \ldots, o_m\}$. The \textbf{\emph{order matching}} mechanism is the process of pairing buy and sell orders within the LOB. Matching typically occurs at fixed intervals (e.g., 500 milliseconds) and follows a priority system based on price and time. Orders are matched sequentially, prioritizing the best price (highest bid or lowest ask) and, for orders at the same price level, earlier submission times according to the time-price priority rule. A trade is automatically executed when a bid price meets or exceeds an ask price in the order book.

\subsection{Market Makers}
In this section, we will introduce the preliminaries of Market Makers (MMers), including the order stacking trading strategy as well as the market indicators used for designing the trading strategy.

\textbf{Order Stacking (OS)} is a trading strategy in Market Making (MM) where market makers (MMers) submit a series of orders ${o_1, o_2, ..., o_n}$ to the \textbf{LOB} $OB_t$ over consecutive time steps ${t_1, t_2, ..., t_n}$. MMers can adjust, cancel, or modify orders proportionally to respond to market dynamics. In MM, \textbf{inventory ($Q$)} refers to the positions held by a market maker after executing trades. To control risk exposure, market makers are subject to an \textbf{inventory holding limit ($d$)}, which defines the maximum long or short position they are allowed to hold for a particular asset at any given time \cite{gurtu2021optimization}. Additionally, MMers typically maintain up to 5 pending orders on each side of the order book to limit market exposure, enable swift position adjustments during volatility, and provide sufficient depth without overcommitting capital.

\textbf{Novel Market Indicators.}  
Beyond traditional market indicators including \textbf{$OHLC$} market prices combining Open, High, Low, Close (\(\mathbf{x}_i\)) and \textbf{Order Book depth (\(D_i\))} measuring the number of buy or sell orders at each price level, RELAVER innovatively incorporates order hold time alongside price and volume as additional factors. We introduce two novel metrics to evaluate order positioning and execution probability: \textbf{Relative Queue Position ($RQP$)} and \textbf{Submitted-order Competitiveness ($C$)}. $RQP$ normalizes an order's position within price-level queues, enabling uniform comparison across different queue depths \cite{parlour1998price}. The comprehensive metric $C$ combines price advantage, relative queue position, and strategic volume dominance to assess order execution likelihood \cite{foucault2005limit}.

\noindent\textbf{Submitted Order Queue.}  
Each order \( o \) in queue \( s_t^s \) is structured as  \( o_t = \) \( (RQP_t^{\tau,i},\ C_t^{\tau},\ P_t,\ q_t,\ t_e,\ t_w,\ \tau) \), with \( t_w \)-based auto-cancellation (\( t' \geq t_e + t_w \)). The queue maintains a capacity of 70 orders, constrained by a 5-order-per-side limit.

\begin{figure*}
    \centering
    \includegraphics[width=0.9\textwidth]{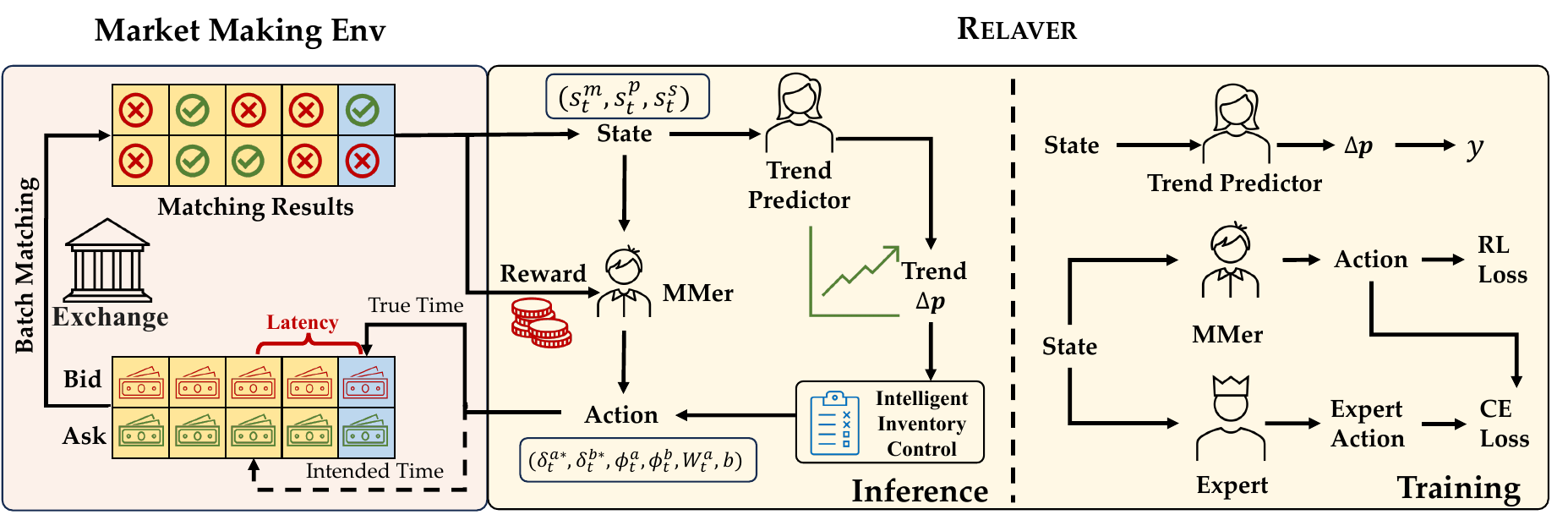}
    \caption{Illustration of the MM environment and \textsc{Relaver}}
    \label{fig:overview}
\end{figure*}

\subsection{MDP Formulation of MM}

Given the preliminaries of the MM and the MMers, we provide the MDP formulate of MM in this section. 

\noindent\textbf{MDP.} A Markov Decision Process (MDP) is represented by the tuple $(S, A, P, r, \gamma, T)$. $S$ and $A$ represent state and action spaces, respectively. $P : S \times A \times S \rightarrow [0, 1]$ is the transition function; $r : S \times A \times S \rightarrow \mathbb{R}$ is the reward function; $\gamma \in (0, 1]$ is the discount factor; and $T$ is the time horizon. The market maker follows a policy $\pi_\theta : S \times A \rightarrow [0, 1]$, aiming to maximize expected cumulative discounted rewards. At each step, the agent observes state $s_t$, takes action $a_t$, receives reward $r_t$, and transitions to $s_{t+1}$. The goal is to find an optimal policy $\pi^*$ that maximizes:
$\mathbb{E}\left[\sum_{t=0}^T \gamma^t r_t \mid s_0\right]$, where $s_0$ is the initial state of the MDP.

\noindent\textbf{State Space $S$.} The state vector $s_t = (s_t^m, s_t^p,s_t^s)$ integrates market microstructure data, portfolio status, and order book dynamics: $s_t^m$ captures 16 market indicators including $OHLC$, $RQP$, $C$, and etc; $s_t^p$ tracks portfolio parameters such as inventory levels, cash reserves, net asset value, and executed trading volume; $s_t^s$ maintains real-time order queue data (positions, quantities, prices) to model pending order interactions with market liquidity.

\noindent\textbf{Action Space $A$.} The agent controls six-dimensional actions: $\delta_t^{a*},\delta_t^{b*}$ regulate quote spreads within ±0.06 of best prices, $\phi_t^{\text{a}},\phi_t^{\text{b}}$ set quoted volumes (0-300 units), and $W_t^a,W_t^b$ define order duration limits (0-5 time units), with orders exceeding wait times being progressively withdrawn through subsequent market transactions.
\begin{equation}
    \mathbf{a}_t = (\delta_t^{a*}, \delta_t^{b*}, \phi_t^{\text{a}}, \phi_t^{\text{b}},W_t^a, W_t^b)
\end{equation}

\noindent\textbf{Reward Function $R$}.
The decision-making mechanism of MMers involves a complex interplay of multiple factors, including \textbf{Profit and Loss (PnL)} represents earnings through bid-ask spreads and mark-to-market valuations; \textbf{Inventory Penalty (IP)} imposes costs when positions deviate from target levels using quadratic functions; \textbf{Exchange Compensation (C)} includes fee rebates and incentives from exchanges to maintain market liquidity ~\cite{niu2023imm}. 

Since \textsc{Relaver} employed an order stacking strategy to optimize our MM approach, we introduced an additional risk measure named \textbf{stacking execution risk (ER)} to account for the potential risks associated with orders remaining in the order book for extended periods. \textbf{Stacking Execution Risk} represents the cumulative risk of unfilled orders in a queue, increasing with order duration and potentially leading to adverse price movements and market impact.
$
ER_t = \sum\nolimits_{i=1}^{n} \left( \sigma \cdot V_i \cdot \left(1 + \frac{t_i}{t_w} \right) \right)
$
where $\sigma$ is the volatility of the market midprice over the last 20 steps and $t_i$ is the time elapsed for the $i_th$ order in the queue.

The reward of the agent is a composition of these terms \[
\mathcal{R}(s_t, a_t, s_{t+1}) = PnL_t - IP_t + C_t - ER_t,
\] which provides a balance between the profit, loss and risks.

\begin{table}[ht]
\centering
\caption{Comparision of previous MM env and our MM env}
\label{tab:compare_env}
\begin{tabular}{c|c|c}
\toprule
     & Previous Env & Our Env \\
\midrule
Latency     & 0 & (30 ms, 80 ms)\\
Matching & Order-by-Order & Batch Matching\\
Cancellation & Instant &  Scheduled\\
\bottomrule
\end{tabular}
\end{table}

\textbf{Advantages of Our MM Env.}
As shown in Table~\ref{tab:compare_env2}, our MM environment enhances realism through three features: (1) introducing random latency (30-80 ms) to reflect actual market response times, replacing zero-latency assumptions; (2) implementing a batch matching system instead of simple Order-By-Order (OBO) approach; and (3) adopting a scheduled cancel mechanism with price-time priority matching, rather than instant cancellations. Last feature potentially reduces cancellation frequency while increasing trading opportunities by allowing orders to remain active for matching with incoming orders. These features collectively improve market stability and fairness: latency simulation prevents unrealistic perfect timing strategies, batch matching reduces the impact of high-frequency arbitrage, and scheduled cancellation promotes more natural order flow and price discovery.

\section{\textsc{Relaver}}

This section introduces the \textsc{Relaver} framework, encompassing three key components. First, the Augmented State-Action Representation details the agent's transition dynamics, including decisions on order pricing, volume, width, and multi-level stacking. Second, the Efficient Q-Teacher guides the agent towards optimal actions by evaluating action quality, thereby enhancing decision-making over time. Third, the Trend Prediction Expert demonstrates intelligent inventory management. We conclude with a comprehensive overview of the \textsc{Relaver} learning procedure, illustrating how these components integrate to form the complete framework.

\subsection{Augmented State-Action Representation}

\begin{figure}[ht]
\centering
\subfigure[$a$ = (2, 4, 1, 3, 2 ,2)]{
\includegraphics[width=0.47\columnwidth]{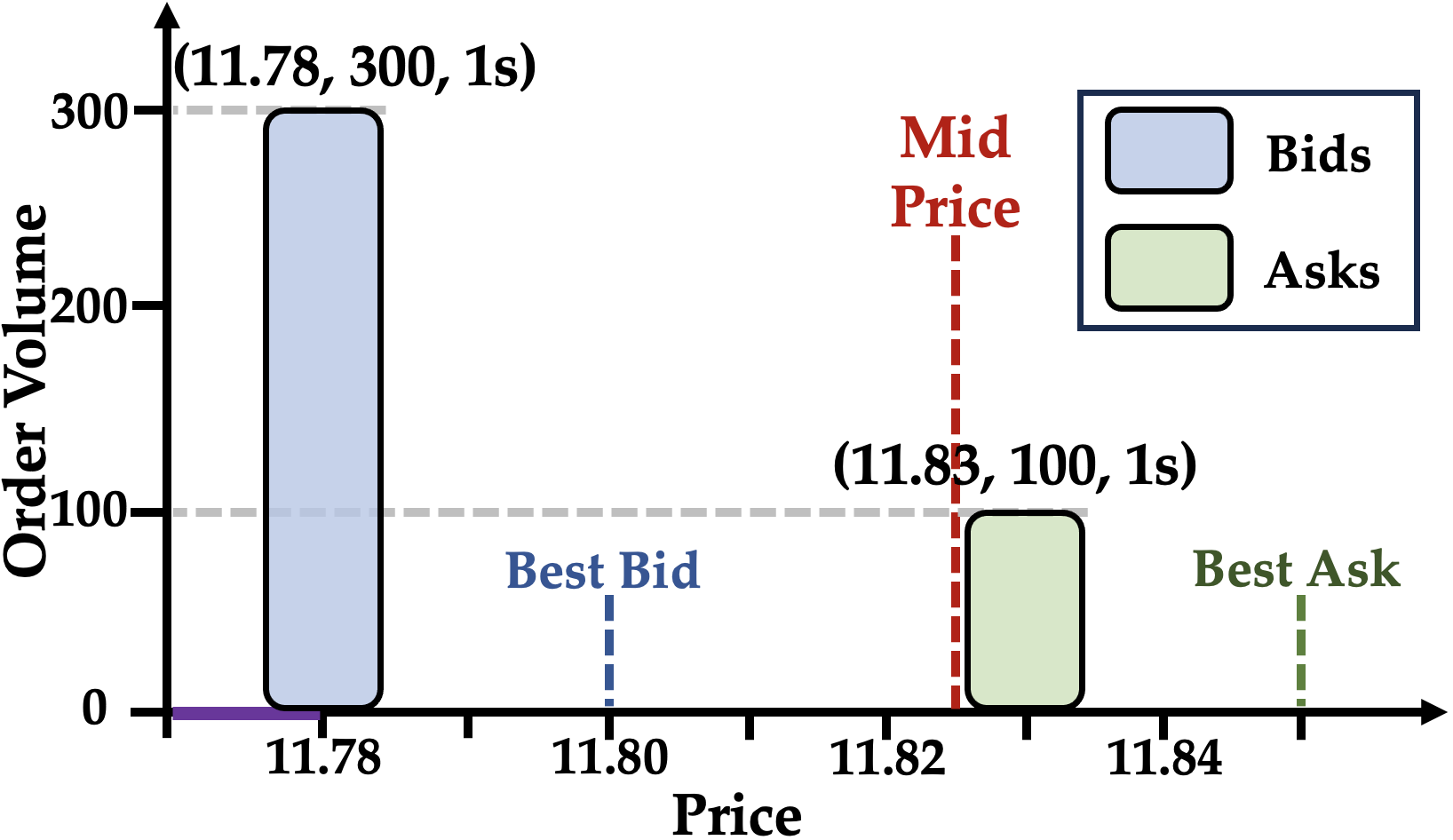}
}
\subfigure[$a$ = (5, 2, 1, 2, 2 ,3)]{
 \includegraphics[width=0.47\columnwidth]{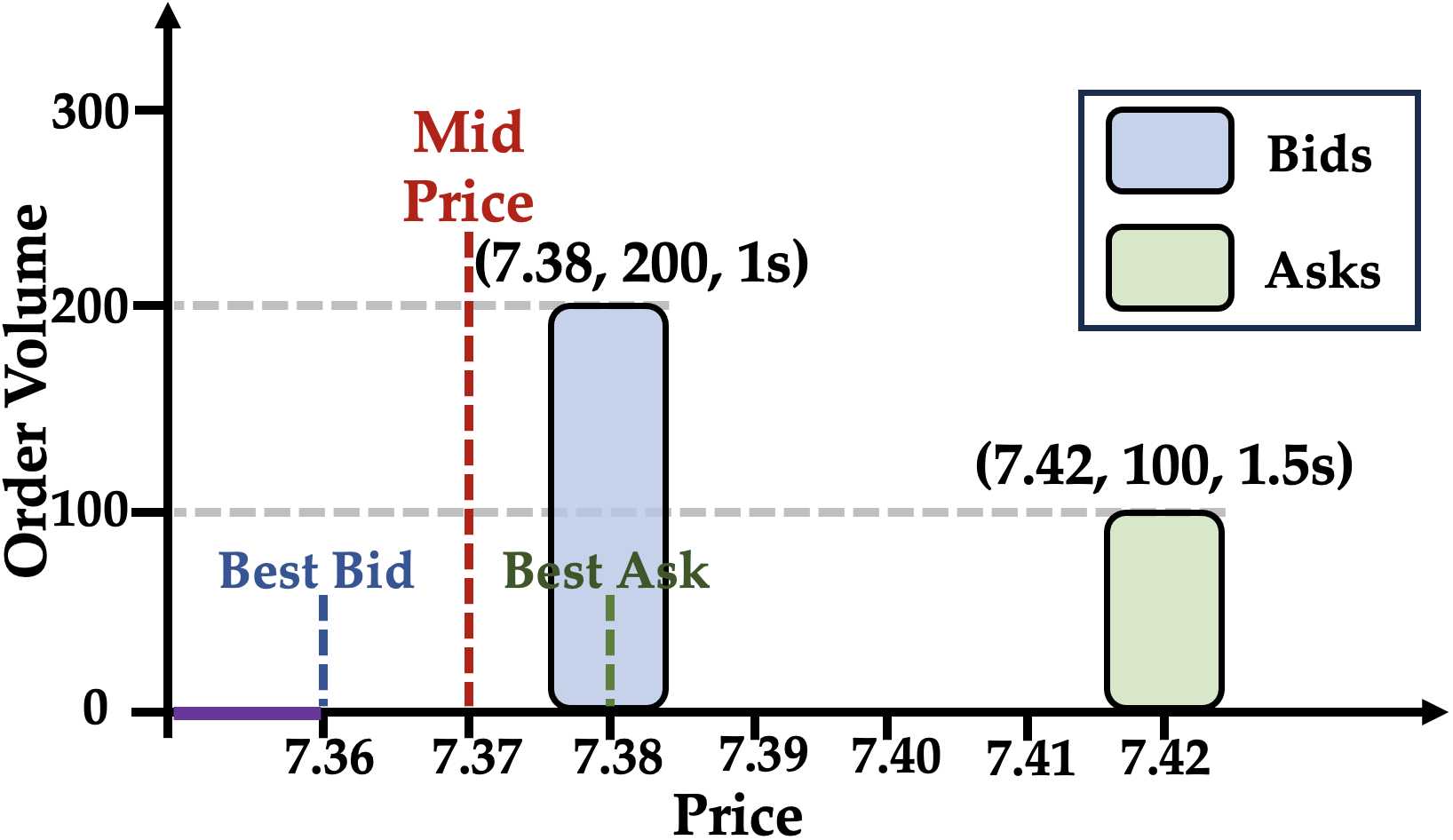}
}
\caption{Illustration of action mapping in the limit order book. Two scenarios demonstrate how actions map to concrete quote placements, where orders are placed relative to the best bid and ask prices. Each scenario shows the resulting quotes with their specific prices, volumes, and durations derived from the action vector.}
\label{fig:action_ill}
\end{figure}

In this section, we will introduce the policy representation for the efficient training. Given the current state $s_{t}$, we introduce the multi-layer perceptron (MLP) to extract the relavent information. However, the current state may not fully reflect the market information, which leads to the suboptimal action. Therefore, we leverage the Long Short-Term Memory (LSTM)~\cite{hochreiter1997long} architecture to model the temporal information from the previous states, i.e., $h_t = \text{LSTM}(\mathbf{s}_t, h_{t-1})$. We introduce an MLP conditional on $h_t$ to get the actions, i.e., $a_{t}=\text{MLP}(h_{t})$. 

The agent's action space is more complex than typical RL scenarios, requiring efficient parameterization. The action is encoded as a six-dimensional discrete vector $(\delta_t^{a*}, \delta_t^{b*}, \phi_t^{\text{a}}, \phi_t^{\text{b}},W_t^a, W_t^b)$, where each dimension corresponds to a specific aspect of the market-making strategy: $\delta_t^{a*}, \delta_t^{b*}$ control price adjustments within $[-0.06, 0.06]$ with step 0.02, $\phi_t^{\text{a}}, \phi_t^{\text{b}}$ determine order volumes within $[100, 300]$ with step 100, and $W_t^a, W_t^b$ set order holding times within $[0.5, 2.5]$ with step 0.5, enabling nuanced control over quote prices, quantities, and order durations. An instance of diverse dual-sided quotes is illustrated in Figure ~\ref{fig:action_ill}. By adopting such action formulation, the agent gains the flexibility to determine both the width and asymmetry of the quotes, relative to the best ask for ask quotes and best bid for bid quotes, as well as their holding time in the order book.

\subsection{RL with Efficient Q-Teacher}
In a market-making environment, the inherent complexity and dynamic nature of the market make it challenging for reinforcement learning algorithms to learn effectively. The essence of MM involves continuously balancing supply and demand while managing risk, which requires rapid adaptation to market fluctuations and efficient strategy optimization. Given these characteristics, employing a market-making dynamic programming using future information to find optimal actions as a Q-teacher becomes particularly advantageous. In our dynamic market-making programming framework, the strategy for determining the optimal action focuses on identifying local optima, due to the inherent complexity of the market-making environment, which renders the discovery of a global optimum infeasible. Within the context of 10 distinct states, our objective is to maximize the total value while minimizing inventory holding costs.
The objective function is defined as
\[
\max_{a_1, a_2, \ldots, a_{10}} \left( \sum\nolimits_{i=1}^{10} V(s_i, a_i) - \lambda \sum\nolimits_{i=1}^{10} C(s_i, a_i) \right),
\]
where \( V(s_i, a_i) \) represents the value by choosing action \( a_i \) in state \( s_i \), \( C(s_i, a_i) \) represents the inventory holding cost incurred by choosing action \( a_i \) in state \( s_i \), and \(\lambda\) is a coefficient balancing trading value and inventory costs.

\begin{algorithm}
\caption{Q-Table Construction}
\label{alg:q_dp}
\begin{algorithmic}[1] 
\STATE \textbf{Input:} Multivariate Time Series $\mathcal{D}$ with Length $N$, Commission Fee Rate $\delta$, Action Space $\mathcal{A}$
\STATE \textbf{Output:} A Table $Q^*$ Indicating Optimal Action Value at Time $t$, Position $p$, and Action $a$
\STATE Initialize \( Q^* \) with shape \((N, |\mathcal{A}|, |\mathcal{A}|)\) and all elements set to 0
\FOR{$t = N-1$ \textbf{down to} $1$}
    \FOR{$p = 1$ \textbf{to} $|\mathcal{A}|$}
        \FOR{$a = 1$ \textbf{to} $|\mathcal{A}|$}
            \STATE $Q^*[t, p, a] \leftarrow \max_{a'} Q^*[t+1, a, a'] + a \times p^b_{t+1} - (p \times p^b_t + E_t(p-a))$
        \ENDFOR
    \ENDFOR
\ENDFOR
\STATE \textbf{return} $Q^*$
\end{algorithmic}
\end{algorithm}

In this context, the \( Q^* \) table is derived using dynamic programming (DP) techniques, as showed in Algorithm~\ref{alg:q_dp}. Specifically, it evaluates the optimal actions across ten states to determine the best possible strategy. By iteratively updating the action values based on expected future rewards, DP ensures that the agent can adapt to complex market dynamics and make decisions that maximize overall value while managing risks.
The DP approach systematically explores all possible states and actions, updating the \( Q^* \) values to reflect the most beneficial strategies. This process enables the selection of optimal actions in each of the ten states, thus guiding the RL training in a structured and efficient manner.

\textbf{RL Training.} During \textsc{Relaver} training with RL, e.g., PPO, these optimal action values serve as a supervision signal, guiding the agent towards an optimal policy more efficiently.
The \textsc{Relaver} loss function is adjusted to include a KL divergence term, measuring the deviation between the current policy \(\pi_{\theta}(a|s)\) and the optimal policy \(\pi^* (a|s)\) derived from the Q-table:
\[
L(\theta) = L_{\text{RL}} + \alpha KL(\pi_{\theta}(a|s) || \pi^* (a|s)).
\]
The policy loss uses the clipped approach from \textsc{Relaver}, incorporating deviations from the optimal policy, same as PPO:
\[
L_{\text{RL}}(\theta) = \mathbb{E} \left[ \min(r_t(\theta) \hat{A}_t, \text{clip}(r_t(\theta), 1 - \epsilon, 1 + \epsilon) \hat{A}_t) \right],
\]
where $\hat{A}$ is the advantage and $\epsilon$ is the bound of the clip.
By introducing the supervision signal of optimal action values, the approach reduces unnecessary exploration, improving training efficiency and effectiveness. This integration provides a structured method to guide policy optimization, resulting in faster convergence and enhanced performance.

\subsection{Pretrained Trend Prediction Expert} 
While Q-Teacher offers theoretically optimal solutions, it inherently presupposes access to future information. In practical scenarios, the absence of such foresight results in policy deviations from the optimal strategy, and the inherent unpredictability of market dynamics leads to suboptimal inventory behaviors, manifesting as either surplus or deficit inventories.

To address these challenges, we introduce a trend predictor to \textsc{Relaver}, which leverages historical MM data to train a LightGBM model for the prediction of market trend based on the current market state $s_m^t$. Specifically, given the current market state $s_m^t$, the trend prediction expert will predict the market's directional movement over the next 30 steps, i.e., the percentage change in market price $\Delta p$.

The prediction target $y$ is categorized into four distinct classes based on the predicted $\Delta p$: bull when $\Delta p > 0.01$, bear when $\Delta p < -0.01$, steady ascent when $0 < \Delta p \leq 0.01$, and  steady descent when $-0.01 < \Delta p \leq 0$.

\begin{figure}[ht]
\centering
\includegraphics[width=0.9\linewidth]{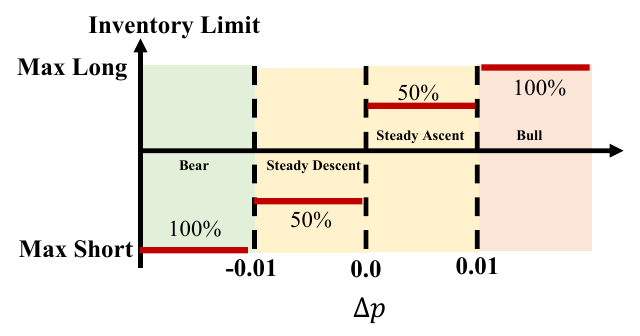}
\vspace{-10pt}
\caption{Market trends and corresponding inventory limits}
\label{fig:trend}
\end{figure}

The inventory limits are dynamically adjusted based on trend classifications to optimize trading positions (Figure~\ref{fig:trend}). Strong trends warrant full exposure (100\%) to maximize profits, while less definitive trends maintain reduced exposure (50\%) for risk management. When inventory positions misalign with predicted trends or exceed limits, market orders are used for swift liquidation, ensuring both risk control and market responsiveness while maintaining capital efficiency.

\begin{table*}[ht]
\centering
\setlength{\tabcolsep}{0.9mm}

\caption{The overall comparison results of \textsc{Relaver} and baselines}
\label{tab:table_overall_pre}
\resizebox{0.95\textwidth}{!}{
\begin{tabular}{c|l|lllll|l}

     \toprule
   &                          & FOIC                   & LIIC                                                         & AS  & $\textsc{DRL}_{\text{os}}$                   & PPO-LSTM                                                & \textsc{Relaver}                                                   \\ 
\midrule
\multirow{3}*{IC} & EPnL[$10^3$]$\uparrow$   & 36.590 ±1.131    & 9.243 ± 0.223                                         & 58.175 ± 0.148 & 70.053 ± 0.308 & 99.461 ± 3.945                                       & \textcolor{blue}{\textbf{109.168}} ± 1.124           \\
   & MAP[unit]$\downarrow$    & 387.584 ± 58.091 & 254.613 ± 55.491                                       & 342.947 ± 91.788 & 117.628 ± 30.635 & 378.374 ± 9.753                                     & \textcolor{blue}{\textbf{\textbf{81.439}}} ± 36.731   \\
   & PnLMAP[$10^3$]$\uparrow$ & 0.0944 ± 0.001   & 0.037 ± 0.004                                          & 0.170 ± 0.003  & 0.625 ± 0.149   & 0.251 ± 0.275                   & \textcolor{blue}{\textbf{\textbf{1.966}}} ± 1.017     \\ 
\midrule
\multirow{3}*{IH} & EPnL[$10^3$]$\uparrow$   & -4.284 ± 0.141    & 9.276 ± 0.293                                         & 1.988 ± 0.374 & 15.322 ± 0.628   & 3.468 ± 1.887                                       & \textcolor{blue}{\textbf{\textbf{89.207}}} ± 2.257      \\
   & MAP[unit]$\downarrow$    & 363.693 ± 69.380 & 247.805 ± 48.493                                       & 401.676 ± 113.540 & 136.095 ± 66.473 & 211.543 ± 122.938                                   & \textcolor{blue}{\textbf{\textbf{105.466}}} ± 70.711  \\
   & PnLMAP[$10^3$]$\uparrow$ & -0.012 ± 0.004    & -0.037 ± 0.002                                         & 0.005 ± 0.001   & 0.134 ± 0.069  & 0.034\textbf{\textbf{~}} ± 0.018                    & \textcolor{blue}{\textbf{\textbf{1.727}}} ± 0.974      \\ 
\midrule
\multirow{3}*{IF} & EPnL[$10^3$]$\uparrow$   & 63.952 ± 0.006    & 27.949 ± 0.011                                          & 44.799 ± 0.511 & 34.080 ± 29.911  & 43.056 ± 0.021                                      & \textcolor{blue}{\textbf{\textbf{93.282}}} ± 39.940   \\
   & MAP[unit]$\downarrow$    & 92.599 ± 48.933   & 360.340 ± 35.469~                                       & 363.274 ± 92.88 & 114.144 ± 91.733 & \textcolor{blue}{\textbf{\textbf{89.600}}} ± 70.246 & 125.494 ± 51.884                                      \\
   & PnLMAP[$10^3$]$\uparrow$ & 0.691 ± 0.078    & 0.078 ± 0.006                                           & 0.124 ± 0.003  & 0.298 ± 0.107  & 0.481 ± 0.005                                       & \textcolor{blue}{\textbf{\textbf{0.743}}} ± 0.003     \\ 
\midrule
\multirow{3}*{IM} & EPnL[$10^3$]$\uparrow$   & 1.834 ± 0.306    & -23.414 ± 16.148                                       & -19.829 ± 11.341   & 5.216 ± 3.066 & 4.023 ± 1.573                                   & \textcolor{blue}{\textbf{\textbf{11.877}}} ± 1.305 \\
   & MAP[unit]$\downarrow$    &  160.557 ± 91.788 & \textcolor{blue}{\textbf{\textbf{77.222}}} ± 113.542 & 103.813 ± 87.122 & 108.330 ± 44.752 & 363.260 ± 20.429                                     & 111.682 ± 21.698                                      \\
   & PnLMAP[$10^3$]$\uparrow$ & 0.011 ± 0.033      & 0.303 ± 0.023                                         & -0.191 ± 0.067 & 0.056 ± 0.042  & 0.011 ± 0.005                                       &  \textcolor{blue}{\textbf{\textbf{0.109}}} ± 0.026   \\
\bottomrule
\end{tabular}
}
\end{table*}

\section{Experiments}
In this section, we present the experiment setup and the experiment results for evaluating the performance of \textsc{Relaver}.
\subsection{Experiment Setup}
\textbf{Market Environments.} Our experiments use historical data from four major Chinese stock index options IC (CSI 500), IF (CSI 300), IH (SSE 50), and IM (CSI 1000), including 5-depth LOB, trade information, and tick data at 500ms intervals. The data spans multiple periods: June-September 2022 (85 days) for trend prediction expert training, October-December 2022 (60 days) for testing, January-July 2023 (138 days) for \textsc{Relaver} model training, and August-October 2023 (59 days) for out-of-sample testing. Trading occurs during Shanghai Futures Exchange hours (9:30-11:30, 13:00-15:00), with 4-hour episodes, 500ms action intervals, and an inventory limit $d$ of ±800 units (8 contracts). Five random seeds were used for robust comparison.

\textbf{Baselines.}
We compared \textsc{Relaver} with three rule-based benchmarks and two state-of-the-art RL approach. 
\begin{itemize}
    \item \textbf{AS}  dynamically adjusts the spread and targets inventory levels based on market volatility, ensuring adequate liquidity provision while optimizing profits through strategic pricing and risk management \cite{avellaneda2008high}.
    \item \textbf{FOIC}  posts bid (ask) orders at the current best bid (ask) while adhering to the inventory constraint $d$ \cite{gueant2013dealing}.
    \item \textbf{LIIC} adjusts its quote prices based on its inventory \( Q \)  \cite{gueant2013dealing}.
    \item \textbf{$\textsc{DRL}_{\text{os}}$ } develops a DRL model for mm under an order stacking framework, using a modified state representation to encode the queue positions of the limit orders\cite{chung2022market}.
    \item \textbf{PPO LSTM} is the PPO algorithm combined with the LSTM neural network that is used to extract temporal patterns from market data \cite{lin2021end}.
\end{itemize}

\textbf{Evaluation Measures.}
We utilize three financial metrics to evaluate the performance of \textsc{Relaver} market-making strategy.

\begin{itemize}
    \item \textbf{Episodic PnL} serves as an optimal metric for evaluating the profitability of an MM agent: $EPnL_T = \sum_{t=1}^{T} PnL_t.$

    \item \textbf{Mean Absolute Position (MAP)} evluate the inventory risk, defined as:
    $MAP_T = \frac{\sum_{t=1}^{T} |z_t|}{\sum_{t=1}^{T} 1 \cdot \mathbb{I}(|z_t| > 0)}.$

    \item \textbf{PnL-to-MAP Ratio (PnLMAP)} integrates the assessment of profitability with the evaluation of inventory risk inherent in a market-making strategy:
    $
    PnLMAP_T = \frac{EPnL_T}{MAP_T}.
    $
\end{itemize}

\textbf{Research Questions.} We aim to address the following three Research Questions (RQs), which are as follows: \textbf{RQ1:} How effectively does \textsc{Relaver} enhance overall market-making performance? \textbf{RQ2:} What are the roles and impacts of the efficient Q-teacher and the pretrained trend prediction expert within the \textsc{Relaver} framework? \textbf{RQ3:} How do various strategies influence the effectiveness of the Efficient Q-Teacher in \textsc{Relaver}?

\subsection{Experiment Results}

\begin{figure*}[ht]
\centering
\subfigure[Up Trend Performance]{
\includegraphics[width=0.92\columnwidth]{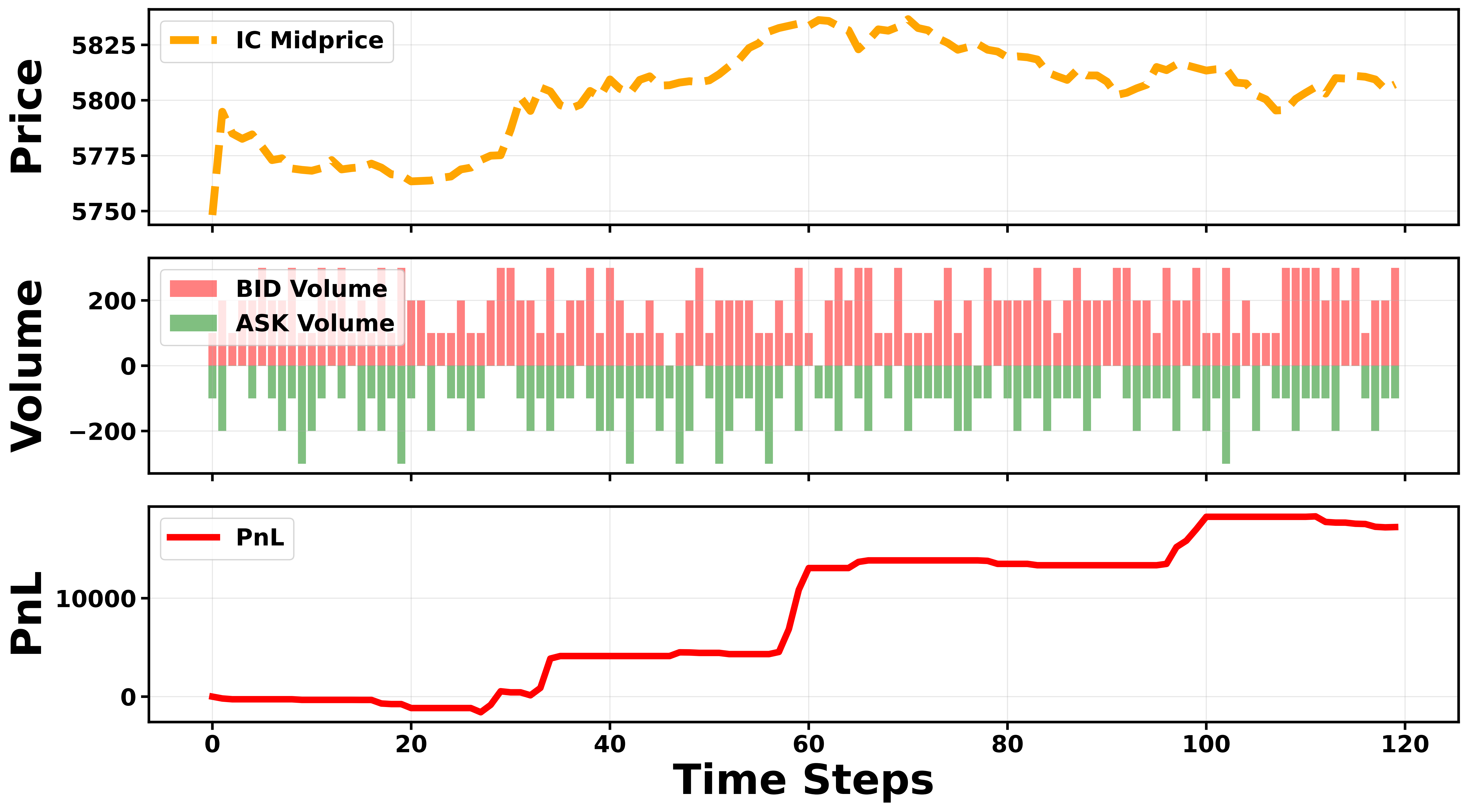}
\label{fig:up}
}
\subfigure[Down Trend Performance]{
 \includegraphics[width=0.92\columnwidth]{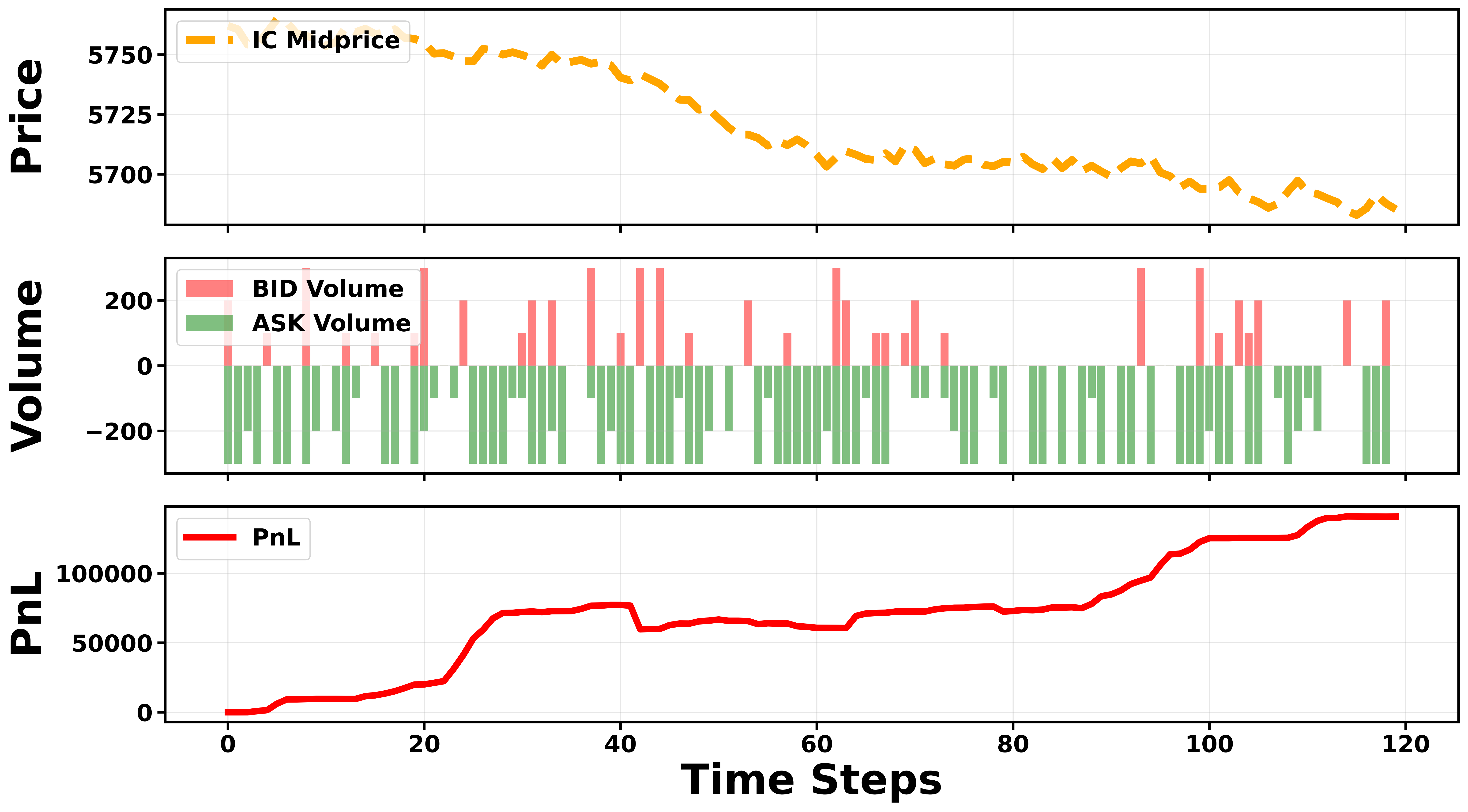}
 \label{fig:down}
}
\vspace{-10pt}
\caption{Up and Down Trend Performance}
\label{fig:interpret1}
\end{figure*}
\textbf{\textbf{RQ1:} Overall MM Performance Enhancement.}
Our initial assessment focuses on determining if \textsc{Relaver} enhances the MM's performance across multiple datasets. The results are shown in Table \ref{tab:table_overall_pre}, which shows that our approach significantly increases overall profit while mitigating inventory risk. 

\textsc{Relaver} consistently achieves the highest expected profit in all market environments, significantly surpassing other methods. Notably, it reaches profits of 109.168 and 89.207 in the IC and IH environments, respectively, and maintains a positive return of 11.877 in the volatile IM environment where others show negative values. This shows its robust profitability and adaptability, which are crucial attributes for effective market-making strategies. Furthermore, \textsc{Relaver} achieves higher profits by holding more inventory. For example, in the IF environment, its MAP is 111.682, higher than that of LIIC and $\textsc{DRL}_{\text{os}}$. Using market trends effectively, it achieves the highest PnL/MAP, demonstrating excellent capital utilization efficiency, which is particularly advantageous for market makers. Furthermore, it maintains a high profit per unit of average position across all environments, with PnL/MAP reaching 1.966 and 1.727 in the IC and IH environments, respectively, far exceeding other methods. This implies that \textsc{Relaver} not only generates substantial profits, but also effectively controls risk, making it an extremely attractive market-making strategy, especially in high-frequency trading where balancing profitability and risk is crucial. Its consistent performance in different market conditions underscores its robustness and versatility for market making.

\textbf{Ablation Study on Market Extreme Volatility.} To elucidate \textsc{Relaver} performance during significant market volatility with distinct upward and downward trends, we performed an ablation study (Figure~\ref{fig:interpret1}). In the analysis of upward trends, \textsc{Relaver} demonstrated an asymmetric volume distribution with bid volumes consistently maintaining approximately 250 units while ask volumes fluctuated between -150 and -200 units, effectively capturing rising market dynamics through strategic inventory accumulation. During downward trends, \textsc{Relaver} exhibited dynamic adjustment by increasing ask volumes to approximately -250 units while maintaining bid volumes around 200 units, demonstrating effective liquidity provision under varying conditions. \textsc{Relaver}'s profitability and risk management capabilities of \textsc{Relaver} were evident in both scenarios. During the upward trend, as the IC mid-price appreciated from 5750 to 5831 (1.4\% increase), the cumulative PnL increased to approximately 12,500. Similarly, in the downward trend scenario where IC mid-price declined from 5769 to 5698 (1.2\% decrease), \textsc{Relaver} maintained profitable operations with a PnL of around 11,000, validating its robustness and adaptability across diverse market conditions.

\textbf{\textbf{RQ2:} Effectiveness of Q-Teacher and Trend Prediction Expert.}
In this study, we evaluated the efficacy of two key components in the \textsc{Relaver} model: Q-Teacher and trend prediction expert. The experimental design encompassed four configurations: the complete \textsc{Relaver} model (\textsc{Relaver}), \textsc{Relaver} without Q-Teacher ($\textsc{Relaver}_{\text{w/o qt}}$), \textsc{Relaver} without Trend Expert ($\textsc{Relaver}_{\text{w/o te}}$), \textsc{Relaver} without Q-Teacher and Trend Expert ($\textsc{Relaver}_{\text{w/o qt, te}}$). The result is shown in Table \ref{tab:variants}. The Q-Teacher component emerged as a critical factor in enhancing the model's profitability. The complete \textsc{Relaver} model achieved EPnL of 109.168, significantly outperforming the version without Q-Teacher, which recorded an EPnL of 68.158. This substantial improvement, approximately 60\%, underscores the effectiveness of Q-Teacher's dynamic programming-based approach in optimizing action selection and maximizing returns in complex market-making scenarios.

\begin{table}[!t]
\small
\centering
\setlength{\tabcolsep}{1mm}
\caption{Performance of \textsc{Relaver} variants with different components on IC dataset}
\label{tab:variants}

\begin{tabular}{c|ccc} 
\toprule
                               & EPnL[$10^3$]$\uparrow$ & MAP[unit]$\downarrow$ & PnLMAP$\uparrow$    \\ 
\midrule
$\textsc{Relaver}_{\text{w/o qt, te}}$

    & 99.46 ± 3.95  & 378.37 ± 9.75    & 0.25 ± 0.28  \\
  $\textsc{Relaver}_{\text{w/o qt}}$    & 68.16 ± 5.49  & \textcolor{blue}{\textbf{73.75}} ± 36.26    & 1.07 ± 0.49  \\
  $\textsc{Relaver}_{\text{w/o te}}$ & 107.96 ± 3.06     & 243.95 ± 6.36   & 0.31 ± 0.01   \\
  \textsc{Relaver}                      & \textcolor{blue}{\textbf{109.17}} ± 1.12   & 81.44 ± 36.73   & \textcolor{blue}{\textbf{1.97}} ± 1.02  \\
\bottomrule
\end{tabular}
\end{table}

The Trend Expert played a crucial role in inventory risk control, significantly reducing the model's Mean Absolute Percentage (MAP) from 243.952 to 81.439, a decrease of approximately 66\%. Even when lacking either Q-Teacher or Trend Expert, \textsc{Relaver} still outperformed the fully ablated version ($\textsc{Relaver}_{\text{w/o qt, te}}$): the model without Q-Teacher excelled in inventory control and profit risk per unit position, while the version without Trend Expert demonstrated superior performance across all metrics, particularly in EPnL and PnLMAP. This indicates that Q-Teacher and trend prediction serve complementary roles within the model, and their synergy creates a robust system that achieves an optimal balance between profitability and inventory management.

\begin{figure}[ht]
\centering
\subfigure[The adverse selection ratio]{
\includegraphics[width=0.47\columnwidth]{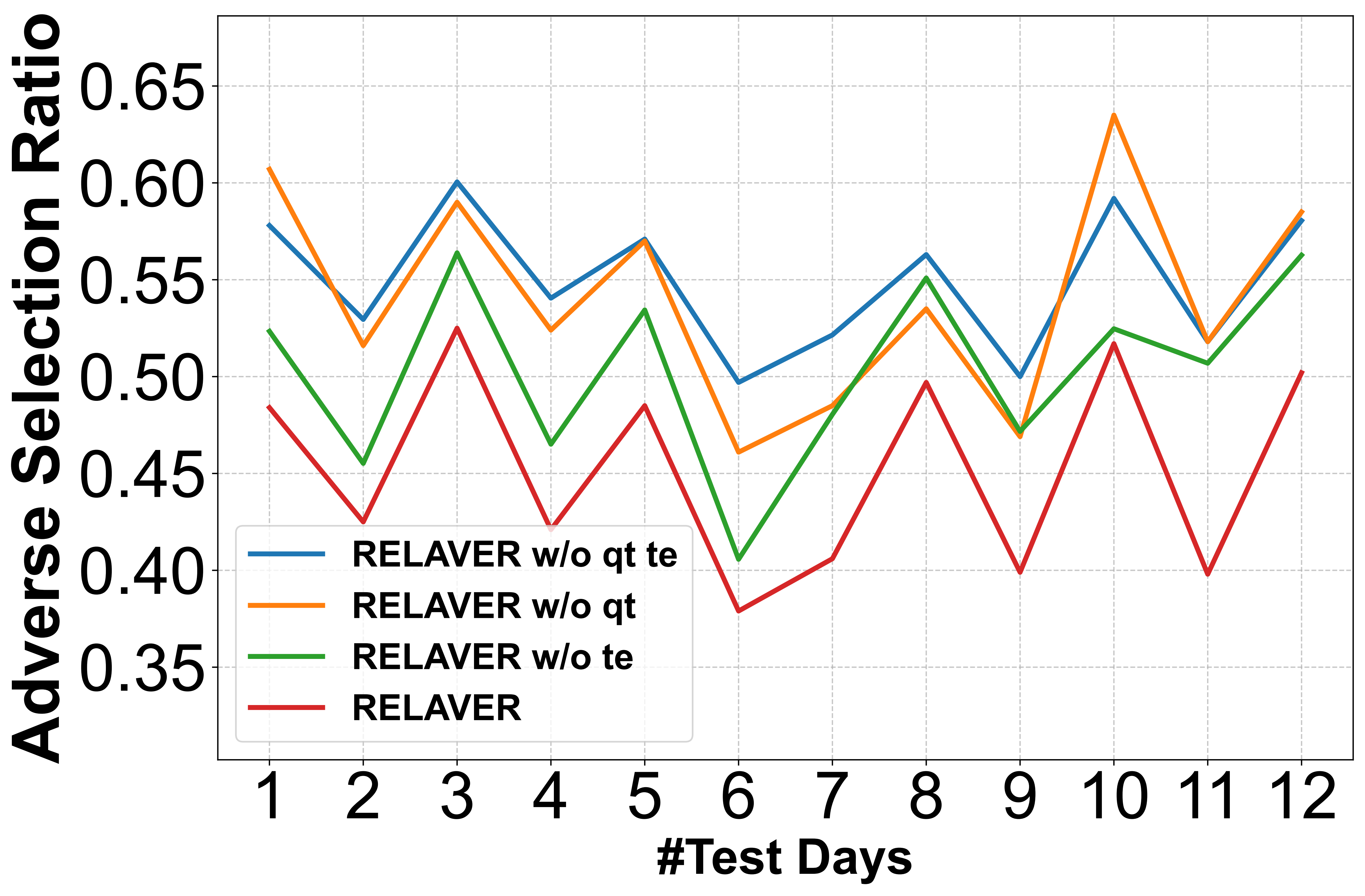}
}
\subfigure[Number of fills]{
 \includegraphics[width=0.47\columnwidth]{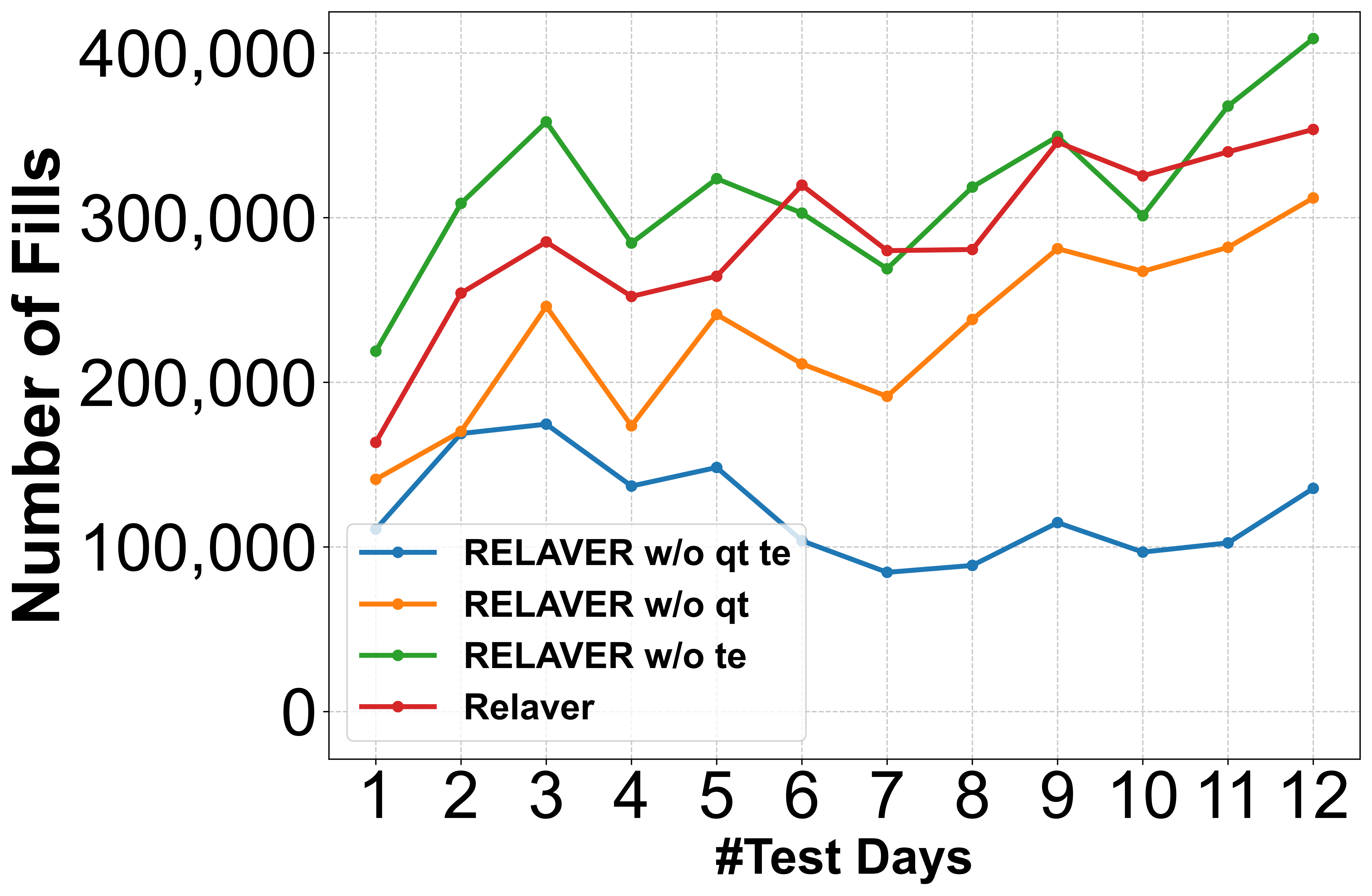}
}
\caption{Testing days of model's daily adverse selection ratio and number of fills}
\label{fig:adv_num}
\end{figure}

For a deeper investigation of the role of \textsc{Relaver} method component, we caculated adverse selection ration, which represents the proportion of trades that result in immediate losses for market best ask (bid) price movement. We also calculated number of fills as the frequency of successful trades executed. The result is shown in Figure~\ref{fig:adv_num}. Based on Figure~\ref{fig:adv_num}(a), it shows that the trend predict expert has great contribution over lowering adverse selection rate. This might because by predicting short-term price movements, it enables the agent to avoid trading against informed counterparties who have superior information about future price directions. Based on Figure~\ref{fig:adv_num}(b), it shows that \textsc{Relaver} and \textsc{Relaver} with Q-Teacher has the greatest fulfillment rate, which demonstrate that Q-Teacher can effectively improve the market-making algorithm's ability to execute trades and meet market demand.

\textbf{\textbf{RQ3:} Effectiveness of different Efficient Q-Teacher}
To investigate the impact of different Q-teacher methods on the performance of the \textsc{Relaver}, we trained and evaluated three rule-based methods as Q-teachers: AS ($\textsc{Relaver}_{\text{w as}}$), FOIC ($\textsc{Relaver}_{\text{w foic}}$), and LIIC ($\textsc{Relaver}_{\text{w liic}}$ ), comparing with two baselines \textsc{Relaver} with DP ($\textsc{Relaver}_{\text{w dp}}$ ) and \textsc{Relaver} without Q-Teacher ($\textsc{Relaver}_{\text{w/o qt}}$) . The result is shown in Table~\ref{tab:QT_rule}. The experimental results demonstrate that for EPnL, \textsc{Relaver} with DP performs best, followed by \textsc{Relaver} with AS and foiic, both greatly outperforming the \textsc{Relaver} without Q-teacher. Notably, the performance of MAP and PnLMAP of these Q-teacher methods varies, our results demonstrate that all variants with Q-teacher implementations outperform the baseline \textsc{Relaver} without Q-teacher. This consistent improvement across different strategies emphasize the potential of Q-teacher methods in enhancing algorithmic trading performance.

\begin{table}[!t]
\centering
\small
\caption{Comparison of \textsc{Relaver} variations with different Q-Teacher rules on IC dataset}
\label{tab:QT_rule}
\begin{tabular}{c|ccc} 
\toprule
                                                & EPnL[$10^3$]$\uparrow$                 & MAP[unit]$\downarrow$                   & PnLMAP$\uparrow$                       \\ 
\midrule
 $\textsc{Relaver}_{\text{w/o qt}}$ & {68.16 ± 5.49}     & {73.75 ± 36.53}     & {1.07 ± 0.49}      \\
 $\textsc{Relaver}_{\text{w as}}$                              & 103.37 ± 14.62                   & 85.76 ± 15.71                     & 1.25 ± 0.34                      \\
  $\textsc{Relaver}_{\text{w foic}}$                             & 86.07 ± 10.73                        & 79.84 ± 14.05                       & 1.17 ~± 0.25                         \\
 $\textsc{Relaver}_{\text{w liic}}$       & 61.18 ± 5.35 & {\textcolor{blue}{\textbf{60.71}} ± 23.44} & {1.12 ± 0.44}  \\
  $\textsc{Relaver}_{\text{w dp}}$                               & \textcolor{blue}{\textbf{109.17}} ± 1.12                   & 81.44 ± 36.73                     & \textcolor{blue}{\textbf{1.97}} ± 1.02                      \\
\bottomrule
\end{tabular}
\end{table}

\section{Conclusion}
In this paper, we propose \textsc{Relaver}, a novel RL-based approach to address latency and inventory risk in MM. First, we formulate an MM environment that incorporates three key realistic features: MM inherent latency, batch matching, and a scheduled order cancellation mechanism. 
Then, \textsc{Relaver} introduce an augmented state-action space that incorporates order hold time alongside traditional price and volume parameters. This enhancement enables our model to optimize execution strategies under latency constraints and time-priority matching mechanisms prevalent in real markets. Sequenctially, \textsc{Relaver} leverage DP as a Q-Teacher to guide the exploration phase during RL training, resulting in more efficient policy learning. Finally, departing from conventional fixed inventory limits,  \textsc{Relaver} incorporates a market trend prediction expert that enables intelligent inventory management, substantially mitigating market risk. Extensive experimental results using four comprehensive historical continuous trading data sets for major Chinese stock index options demonstrate that \textsc{Relaver} significantly outperforms five baseline models in overall MM performance, while validating the effectiveness of each proposed component.

\bibliographystyle{named}
\bibliography{ijcai25}

\clearpage

\appendix

\section{Notations}

\begin{table}[H]
\centering
\resizebox{0.45\textwidth}{!}{
\begin{tabular}{c|p{6cm}}
\toprule
Notation     & Description \\ \midrule
$t$ & Time Step t \\ 
$s^{l}_{t}$ & Latency between order placement and execution  \\ 
$t_{e}$ & Time of order execution \\ 
$o_t$  &  Submitted order at time $t$\\
$p_t$ & Submitted order price at time $t$\\ 
$q_t$ & Submitted order volume at time $t$\\
$\tau$  &  ``ask'' or ``bid'' side of order \\ 
$OB_t$   &  Limit Order Book at time $t$ \\ 
\midrule
$OS$ &  Order Stacking  \\ 
$Q$ & Inventory held by MMers \\ 
$NV$  &  Overall value of MMers portfolio (Net Value) \\ 
$d$  &  Inventory Holding Limit \\ 
$D_i$   &  Order Book Depth at the $i-th$ price level \\ 
$S_i$   &  Bid-Ask Spread at the $i-th$ price level\\ 
$I_i$    &  Imbalance at the $i-th$ price level \\
$OF_i$   &  Order Flow at the $i-th$ price level
\\ \midrule
$OHLC$  &  Opening, highest, lowest and closing prices at time $t$ \\
$RQP_t^{\tau,i}$  &  Submitted Order's relative queue position at price level $i$ \\
${l_{t}^{front,\tau,i}}$  &  Preceding submitted order count at price level $i$, time $t$\\
$l_{t}^{behind,\tau,i}$ & Succeeding submitted order count at price level $i$, time $t$ \\
$C_t^{\tau}$  &  Submitted order competitiveness at time $t$ \\
$P_t^{best,\tau}$   &  The highest ``ask''/``bid'' price in the queue \\
$V_t^{front,\tau,i}$   & Preceding total volume of order at price level $i$, time $t$ \\
$V_t^{total,\tau,i}$  & Total order volume at price level $i$, time $t$ \\ \midrule
$V_t^{\tau,i}$  & Trading volume of current order at price level $i$, time $t$ \\
$s_t^s$  & Submitted order queue at time $t$ \\
$t_w$  & Submitted order waiting time at time $t$\\
$m_t$  &  Market midprice at time $t$ \\
$PnL_t$  &  MMers profit and loss  \\
$IP_t$ & Inventory penalty at time $t$\\
$C_t$ & Exchange Compensation at time $t$\\
$ER_t$ & Total risk associated with unfilled orders at time $t$ \\

\bottomrule
\end{tabular}}
\caption{Notation table}
\label{tab:notation}
\end{table}

\section{Frequently Asked Questions}
In this section, we address potential questions that readers may have through a set of Frequently Asked Questions (FAQs), providing detailed explanations to enhance the clarity of our work.

\textbf{\textbf{Q1: Are the baseline models used for comparison in this article fair?}}
In the field of RL for market making (MM), although there is no unified standard environment, our comparison methodology aligns with the benchmark comparisons established in the latest top-tier article ~\cite{niu2023imm}. With very few papers providing open-source implementations, most existing works have overly simplified environments without considering crucial factors like latency and batch matching, making them difficult to migrate to our realistic setting. As the first work that applies reinforcement learning to market making with these realistic constraints, our benchmark choices reflect the current state of research and provide an important reference standard for future studies.

\textbf{\textbf{Q2: What are the challenges in adapting other RL market makers to your environment?}}
The key challenge in transplanting other RL-based market makers to our environment lies in their fundamental assumptions about market mechanics. While general RL methods like PPO+LSTM can adapt to any environment structure, specialized approaches like IMM are designed for continuous order matching and immediate execution feedback, making them incompatible with our more realistic setting that incorporates latency and batch order matching. These differences in core mechanisms make direct implementation of existing methods impractical without significant modifications that could alter their original characteristics.

\textbf{\textbf{Q3: How does the model address computational efficiency and performance when handling in real-time trading scenarios?}}
The computational concern for high-frequency scenarios is valid. However, \textsc{Relaver} employs offline training. After the model converges, it only requires inference on real-time data, which is computationally efficient. In our experiments, using a single NVIDIA RTX 4070 GPU, the inference latency is consistently between 1.8 and 2.5 ms, which is negligible compared to the overall operational time requirements. The latency of execution during deployment can also be optimized through parallel processing and hardware acceleration. While top-tier firms operate at thousands of trades per second, our approach targets market makers operating at moderate frequencies (0.5s per decision), making it computationally feasible with standard hardware acceleration. 

\textbf{\textbf{Q4: How did you determine the coefficients in your reward function?}}
The reward function coefficients follow the standard formulation of ~\cite{niu2023imm} with one extension: we introduce the risk component of order stack execution. This addition is inspired by industry practice~\cite{chen2019inventory} and specifically evaluates the risk mitigation aspect of our system's order stacking mechanism.

\textbf{\textbf{Q5: What is the rationale for choosing the best price matching rule?}}
We adopt the best price rule (highest bid-lowest ask) as it is the common practice in major exchanges (NYSE, NASDAQ) and in the RL for market making literature ~\cite{xu2022performance,niu2023imm}.
 
\textbf{\textbf{Q6: What is the cost of model retraining when Q-Teacher's performance drops?}}
The Q-teacher's optimal action table is precomputed using dynamic programming, requiring only 6 hours on a single NVIDIA RTX 4070 GPU. Although periodic retraining may be needed, this offline process incurs no additional cost during \textsc{Relaver} training and can be further optimized in production environments.

\textbf{\textbf{Q7: Clarify how your trend prediction system works and integrates with RL?}}
The trend prediction expert enforces risk control by triggering market orders when inventory conflicts with predicted trends. These forced liquidations create transaction costs that train the RL agent to align with market predictions. The LightGBM model leverages current market microstructure data for prediction, as supported by price formation theory. The 30-step (15 second) horizon is optimized to capture market impact and mean reversion cycles.

\textbf{\textbf{Q8: What is your accessibility contribution?}}
We establish a comprehensive standardized simulation environment with real latency and matching algorithms, which will be open source in its entirety. This contribution represents a pivotal advancement in market-making research infrastructure, providing a canonical benchmark environment for the RL-based market-making research community. By facilitating rigorous comparative analyses and enabling reproducible research, we accelerate the field's progression through standardized evaluation protocols. From an industrial perspective, our framework eliminates the binary constraint of microsecond-level infrastructure requirements, expanding market-making accessibility to a broader spectrum of market participants. This creates a more inclusive trading ecosystem where technological sophistication becomes a gradient rather than a barrier. We have made our code available at \url{https://anonymous.4open.science/r/Relaver_ijcai-3025/}. Once accepted, we will provide a more detailed codebase and comprehensive instructions.

\textbf{\textbf{Q9: Is there a relationship between PPO-LSTM and Relaver (without Q-Teacher/trend prediction), given their similar performance?}}
Our \textsc{Relaver}  model is built on PPO-LSTM as the base architecture, which is thoroughly described in Section 4.1. In the ablation studies, when we remove both the Q-Teacher and trend prediction expert components, the model is reduced to the vanilla PPO-LSTM. We explicitly name it as\textsc{Relaver} without Q-Teacher and trend prediction expert in our experiments to maintain consistent terminology throughout the paper, although it is essentially equivalent to the standard PPO-LSTM baseline.
\section{Related Work}
\label{app:related_work}
In this section, we provide a complete review of the related work. The first line is the MM environment modeling. 
Traditional market making (MM) approaches, as exemplified by 
~\cite{ho1981optimal,labadie2013high,gueant2013dealing}, often employ simplified environment models that treat MM as an analytically solvable stochastic optimal control problem. These models use Poisson processes to represent random order arrivals and price changes, focusing on inventory risk and expected utility maximization for market makers, which laid the groundwork for future MM research. For example, 
\cite{guilbaud2013optimal} constructed a high-frequency trading environment where the asset's midprice is assumed to follow an arithmetic Brownian motion, and they used the Hamilton-Jacobi-Bellman equation to solve this continuous-time, finite-horizon stochastic control problem. Following this work, \cite{cartea2015risk} incorporate more sophisticated risk metrics Conditional Value-at-Risk and traders' risk aversion levels, enabling more refined and personalized risk management. However, such models rely on idealized stochastic processes and data distributions rather than empirical market data, undermining their practical applicability in actual market-making environments. Consequently, employing more advanced modeling techniques, such as RL, may offer solutions to more realistic and complex MM scenarios.

In recent years, the field of MM has seen a significant surge in the application of RL techniques ~\cite{kumar2020deep,ganesh2019reinforcement,chung2022market,gavsperov2021market,jerome2022market}. RL approaches in MM can be categorized into two classes. The first, which is predominant, adopts a single-price level strategy ~\cite{sadighian2019deep,gueant2019deep,xu2022performance,beysolow2019market}. Within this class, action spaces are defined either as pre-specified discrete sets, such as an octuple of preset half-spread pairs, or as continuous "half-spread" spaces allowing selection of a continuous half-spread on each side of the order book. However, single price strategies in MM lack the flexibility to adapt to complex market dynamics and fail to effectively manage inventory risks at different price levels. In addition, these strategies miss potential profit opportunities and are more vulnerable to exploitation by other market participants.

Another approach in MM employs multi-price level strategies with more complex action spaces, placing order volumes at multiple price points on both sides of the orderbook~\cite{chakraborty2011market,abernethy2013adaptive}. These complex action spaces allow for simultaneous decisions on prices and volumes across multiple levels, enabling more precise and flexible responses to market dynamics. The latest work \cite{niu2023imm} designed to capture complex multi price level strategies within a low-dimensional representation. It enables sophisticated order stacking strategies while maintaining a manageable action space for RL algorithms, allowing the agent to flexibly determine both the width and asymmetry of quotes relative to a reference price. However, it may lead to frequent order cancellations as the agent adjusts its orders after each non-execution, potentially causing market disruption and increased computational overhead.

Unfortunately, providing more accurate and realistic MM trading decisions requires a multi-price level strategy with a fine-grained action space. This approach involves considering longer-term market fluctuations and preemptively reserving orders in the orderbook to obtain better price margins and increased trading possibilities. Moreover, previous work has overlooked the inherent latency in markets, leading to unrealistic execution assumptions, mispriced risk, and overly optimistic performance estimates \cite{gao2020optimal}. 
\section{Problem Statement}

In this section, we introduce fundamental financial concepts utilized in simulating the market-making process and formulate a more realistic environment as a Markov Decision Process (MDP) framework for MM.

\subsection{Market Making}
        
In this section, we will introduce the preliminaries of Market Making (MM) including the definitions of latency, limited order book and order matching mechanism to facilitate the understanding.

\begin{definition}[Latency]
The \textbf{\emph{latency}} in MM is defined as the time $s^{l}_{t}$ between the order placement and the execution, which is typically modeled as a random variable $\mathcal{U}(\cdot)$, i.e., $s^{l}_{t}\sim \mathcal{U}(\cdot)$. Therefore, the time of order execution $t_{e}=t+s^{l}_{t}$ where $t$ is the time of the order placement.
\end{definition}

\begin{definition}[Order]
The \textbf{\emph{order}} in MM is defined as the action where a MMers submits an order at time $t$ with a specified price $p_t$ and quantity $q_t$. Denoted as a tuple $o_t = (p_t, q_t, t_{e}, \tau)$, where $\tau \in \{\text{ask}, \text{bid}\}$.
\end{definition}

\begin{definition}[Limited Order Book (LOB)]
\label{def:lob}
The \textbf{\emph{Limit Order Book}} (LOB) in MM is defined as the collection of all active orders at a given time $t$, denoted by $OB_t$. 
We denote an m-level LOB at time $t$ as $OB_t = \{o_1, o_2, o_3 \ldots, o_m\}$.

\end{definition}

\begin{definition}[Order Matching]
The \textbf{\emph{order matching}} mechanism refers to the process of pairing buy and sell orders in the Limit Order Book $OB_t$. It typically occurs at intervals of approximately 500 milliseconds and matches orders $o_t \in OB_t$ where $t_0 \leq t$, with $t$ being the auction time. Orders are matched sequentially, prioritizing the best price (highest bid or lowest ask) and earlier submission time for orders at the same price level. A trade occurs when a bid order price meets or exceeds a ask order price.
\end{definition}

{\textbf{\emph{Market making}} is a continuous process of quoting both bid and ask order $o$ for a financial instrument. This process aims to profit from the bid-ask spread while providing liquidity to the market. These orders are updated based on the current state of the Limit Order Book $OB_t$, market conditions, and inventory positions, subject to the latency $s^l_t$ between quote submission and execution. The effectiveness of market making is measured by its ability to maintain a profitable bid-ask spread, provide consistent liquidity, and manage risk in varying market conditions.}

\subsection{Market Makers}
In this section, we will introduce the preliminaries of Market Makers (MMers), including the order stacking trading strategy, the inventory and net values, as well as the market indicators used for designing the trading strategy.

\begin{definition}[Order Stacking (OS)]
\textbf{\emph{Order Stacking}} in MM is defined as a trading strategy where MMers submits a series of orders ${o_1, o_2, ..., o_n}$ to the Limit Order Book (LOB) $OB_t$ over consecutive time steps ${t_1, t_2, ..., t_n}$. This strategy creates a dynamic price ladder in the LOB over time, allowing for gradual order placement and execution as market conditions change. MMers could adjust, cancel, or modify orders proportionally. 
\end{definition}

\noindent\textbf{Maximum Pending Orders.}
MMers typically maintain up to 5 pending orders on each side of the order book to limit market exposure, enable swift position adjustments during volatility, and offer sufficient depth without overcommitting capital.

\noindent\textbf{Inventory and Net Value.}
In MM, inventory ($Q$) refers to the positions held by a market maker after executing trades and net value ($NV$) refers to the overall value of a market maker's inventory after accounting for any liabilities.

\noindent\textbf{Inventory Holding Limit}
In MM, inventory holding limit ($d$) refers to the maximum long or short position of a particular asset that a market maker is allowed to hold in their inventory at any given time \cite{gurtu2021optimization}. This limit is typically set to control risk exposure and maintain market neutrality. 

\noindent\textbf{Normal Market Indicators.}
Fundamental Market Microstructure Indicators are key metrics extracted from the order book, providing insights into market liquidity, depth, and price dynamics:

\begin{itemize}[leftmargin=2em]
    \item \textbf{Order Book Depth (\( D_i \)):} Number of buy/sell orders at the \( i \)-th price level.
    
    \item \textbf{Bid-Ask Spread (\( S_i \)):} \( S_i = P_{\text{ask},i} - P_{\text{bid},i} \), where \( P_{\text{bid},i} \) and \( P_{\text{ask},i} \) are bid and ask prices at the \( i \)-th level.
    
    \item \textbf{Imbalance (\( I_i \)):} \( I_i = Q_{\text{buy},i} - Q_{\text{sell},i} \), where \( Q_{\text{buy},i} \) and \( Q_{\text{sell},i} \) are buy and sell quantities at the \( i \)-th level.

    \item \textbf{Order Flow (\( OF_i \)):} \( OF_i = N_i - C_i + M_i \), where \( N_i \), \( C_i \), and \( M_i \) are new orders, cancellations, and modifications at the \( i \)-th level, reflecting the update of order instantly.
    
    \item \textbf{OHLC:} \( \mathbf{x}_t = (p_t^o, p_t^h, p_t^l, p_t^c) \), where \( p_t^o \), \( p_t^h \), \( p_t^l \), \( p_t^c \) are the opening, highest, lowest and closing prices at time \( t \).

\end{itemize}

\noindent\textbf{Novel Market Indicators.} Using a relative queue position provides a standardized measure, enabling easier comparisons of submitted order positions between price levels ~\cite{parlour1998price,chung2022market}. At time step $t$, the relative queue position of the submitted order at the price level $i$ is defined as: $RQP_t^{\tau,i} = \frac{l_{t}^{front,\tau,i}}{l_{t}^{front,\tau,i} + l_{t}^{behind,\tau,i}}$
, where ${l_{t}^{front,\tau,i}}$ is the order length in front of the submitted order and $l_{t}^{behind,\tau,i}$ is the order length behind the submitted order at the price level $i$. 

We also define submitted-order competitiveness as a novel indicator that measures the relative advantage of the submitted order. This concept, introduced for the first time in the RL for MM, is formulated as follows:
$C_t^{\tau} = \frac{P_t^{\tau}}{P_t^{best,\tau}} \cdot \left(1 - \frac{V_t^{front,\tau,i}}{V_t^{total,\tau,i}}\right) \cdot \frac{V_t^{\tau,i}}{V_t^{avg,\tau,i}}$, where $P_t^{\tau}$ is the submitted price of the current order, $P_t^{best,\tau}$ is the highest ask/bid price in the queue, $V_t^{front,\tau,i}$ is the total order volume in front of the submitted order at $i$, $V_t^{total,\tau,i}$ is the total order volume $V_t^{\tau,i}$ is the trading volume of the current order at $i$, $V_t^{avg,\tau,i}$ is the average order trading volume at $i$ ~\cite{foucault2005limit,harris2002trading}. By analyzing the competitiveness of the bid and ask sides, agent can better understand our order's competitive edge in the market and assess the likelihood of order execution over time.

\noindent\textbf{Submitted Order Queue.}
Each order $o$ in the submitted order queue $s_t^s$ is represented as follows:
\[o = (RQP_t^{\tau,i}, C_t^{\tau}, P_t, q_t, t_e, t_w, \tau),\]
where $t_w$ is the order waiting time. For an order $o$ placed at time $t$, the order is automatically canceled if the next step $t'$ is: $t' \geq t_e + t_w$.
Since there are at most five orders for both sides of the queue that exist simultaneously in the order book, the size of the queue is 70.

\subsection{MDP Formulation of MM}

Given the preliminaries of the MM and the MMers, we provide the MDP formulate of MM in this section. 

\noindent\textbf{MDP.} A Markov Decision Process (MDP) is represented by the tuple $(S, A, P, r, \gamma, T)$. $S$ and $A$ represent state and action spaces, respectively. $P : S \times A \times S \rightarrow [0, 1]$ is the transition function; $r : S \times A \times S \rightarrow \mathbb{R}$ is the reward function; $\gamma \in (0, 1]$ is the discount factor; and $T$ is the time horizon. The market maker follows a policy $\pi_\theta : S \times A \rightarrow [0, 1]$, aiming to maximize expected cumulative discounted rewards. At each step, the agent observes state $s_t$, takes action $a_t$, receives reward $r_t$, and transitions to $s_{t+1}$. The goal is to find an optimal policy $\pi^*$ that maximizes:
$\mathbb{E}\left[\sum_{t=0}^T \gamma^t r_t \mid s_0\right]$, where $s_0$ is the initial state of the MDP.

\noindent\textbf{State Space $S$.} The state at step $t$ is formed with three parts $s_t = (s_t^m, s_t^p,s_t^s)$. Specifically, $s_{t}^{m}$ denotes the 16 market indicators, including OHLV, top 3 levels $i$ of imbalance ($I_i$), order flow ($OF_i$), Best bid price ($P_{\text{bid},i}$), Best ask price ($P_{\text{ask},i}$). $s_t^p$ denotes the private information from the portfolio, including the inventory, remaining case, net value, total traded volume executed of the portfolio. $s_t^s$ is the queue of orders submitted, including queue positions, order quantities and prices. We note that the queue of orders submitted $s_t^s$ can help the agent track orders and capture the complex dynamics of the market, thus helping the agent make accurate predictions and efficient decisions.

\noindent\textbf{Action Space $A$.}
$\delta_t^{a*}$ denotes the desired ask spread from the best ask price, with a range of ±0.06, while $\delta_t^{b*}$ denotes the desired bid spread from the best bid price, also with a range of ±0.06. The variable $\phi_t^{\text{a}}$ represents the desired quoted ask volume, with a range of 300 units, and $\phi_t^{\text{b}}$ represents the desired bid volume, also quoted within a range of 300 units. The term $W_t^a$ indicates the associated wait time for the ask order, with a maximum duration of 5. Similarly, $W_t^b$ represents the associated wait time for the bid order. If this holding time is exceeded, the orders in the order book will be automatically and uniformly withdrawn as subsequent trades occur.
\begin{equation}
    \mathbf{a}_t = (\delta_t^{a*}, \delta_t^{b*}, \phi_t^{\text{a}}, \phi_t^{\text{b}},W_t^a, W_t^b)
\end{equation}

\noindent\textbf{Reward Function $R$}.
The decision-making mechanism of MMers involves a complex interplay of multiple factors, including profit and loss, inventory risk and compensation from the exchange. Since \textsc{Relaver} employed an order stacking strategy to optimize our MM approach, we introduced an additional risk measure named stacking execution risk to account for the potential risks associated with orders remaining in the order book for extended periods.
\begin{itemize}[leftmargin=2em]
    \item \textbf{Profit and Loss (PnL)} measures the MMers' financial performance by tracking the difference between the revenue from executed trades and the costs incurred. The $PnL$ at time step $t$ is defined as
$PnL_t  = \left( \sum\nolimits_{i \in A_t} p_i^a \cdot v_i^a - \sum\nolimits_{j \in B_t} p_j^b \cdot v_j^b \right) \\+ (m_{t+1} - m_t) \cdot Q_{t+1}
$,
where \( m_t\) denotes the market midprice at time step $t$; \( p^a, v^a, p^b, v^b \) represent the price and volume of the filled ask (bid) orders respectively; \( Q \) signifies the current inventory, with \( Q > 0 \) when the agent holds a longer position.
\item \textbf{Inventory Penalty}. To manage inventory risk while allowing for strategic inventory holding, we implement an inventory damping mechanism to penalize the holding: 
$
IP_t = -\eta |Q_t| \cdot \mathbb{I}(|Q_t| > d),
$ where $\eta$ is the coefficient.
A penalty for inventory holding is applied only when the inventory $Q_t$ exceeds inventory limit $d$.

\item \textbf{Exchange Compensation} is the incentives or rebates provided by exchanges to MMers for providing liquidity, encouraging active participation in maintaining market efficiency:
$
C_t = \beta \left( \sum\nolimits_{i \in A_t} p_i^a \cdot v_i^a + \sum\nolimits_{j \in B_t} p_j^b \cdot v_j^b \right).
$

\item \textbf{Stacking Execution Risk} denotes the cumulative risk associated with a queue of unfilled orders, which increases as orders remain unexecuted over time, potentially exposing the trader to adverse price movements and increased market impact:
$
ER_t = \sum\nolimits_{i=1}^{n} \left( \sigma \cdot V_i \cdot \left(1 + \frac{t_i}{t_w} \right) \right)
$
where $\sigma$ is the volatility of the market midprice over the last 20 steps and $t_i$ is the time elapsed for the $i_th$ order in the queue.
\end{itemize}

The reward of the agent is a composition of these terms \[
\mathcal{R}(s_t, a_t, s_{t+1}) = PnL_t + IP_t + C_t + ER_t,
\] which provides a balance between the profit, loss and risks.

\begin{table}[ht]
\centering
\caption{Comparision of previous MM env and our MM env}
\label{tab:compare_env2}
\begin{tabular}{c|c|c}
\toprule
     & Previous Env & Our Env \\
\midrule
Latency     & 0 & (30 ms, 80 ms)\\
Matching & Order-by-Order & Batch Matching\\
Cancellation & Instant &  Scheduled\\
\bottomrule
\end{tabular}
\end{table}

\textbf{Advantages of Our MM Env.}
Our MM environment significantly improves upon previous models by incorporating three key realistic features, in contrast to the simplified assumptions of earlier simulations, as shown in Table ~\ref{tab:compare_env2}. We introduce a random (30 ms, 80 ms) latency to simulate actual market response times, replacing the unrealistic zero-latency assumption. Our batch matching system better reflects complex price determination processes, advancing beyond the simplistic Order-By-Order (OBO) approach. Additionally, we implement a scheduled cancel mechanism mirroring real exchange practices, replacing the instant cancel/modify option used previously. This change, combined with a price-time priority matching mechanism, may decrease order cancellation frequency while potentially increasing trading opportunities. By allowing orders to remain in the market for a set period, they have a chance to match with incoming orders based on their price and time of entry, potentially executing trades that would miss with instant cancellations. 
\section{More Introduction of MM}
\textbf{Latency.}
A key aspect of market making is the inherent latency between placing and executing an order. This latency, denoted as $s^l_t$, can vary and is often modeled as a random distribution to reflect real-world conditions. For example, it might be assumed to fall within a range such as \( (30 \text{ ms}, 80 \text{ ms}) \). This range is illustrative, capturing typical delays due to network transmission times, processing times, and other operational factors that can vary based on technology infrastructure and market conditions.
This means that if an order is placed at time \( t \), the actual execution time \( t_{\text{e}} \) is calculated as:
$t_{\text{e}} = t + s^l_t$

This random latency is crucial for simulating the inherent delays in information transmission and order execution found in real markets. These delays reflect the time required to receive market data and process orders, impacting the efficiency of market operations.

\textbf{Order Matching.}
Market making involves the continuous provision of buy and sell prices to facilitate trading and provide liquidity. A critical component of this process is the matching mechanism, which operates based on time and price priority principles, often incorporating auction-like features. The price priority rule ensures that orders with the best price are executed first. For buy orders, this means higher prices are prioritized: if $P_{\text{buy1}} > P_{\text{buy2}}$, then buy1 is matched first. For sell orders, lower prices take precedence: if $P_{\text{sell1}} < P_{\text{sell2}}$, then sell1 is matched first. This rule incentivizes traders to offer better prices, thereby improving market efficiency and liquidity. When multiple orders have the same price, the time priority rule is applied, dictating that the order placed first will be executed first: if $t_{\text{eO1}} < t_{\text{eO2}}$ and $P_{\text{order1}} = P_{\text{order2}}$, then order1 is matched first. Time priority encourages traders to act quickly and contributes to a fair and orderly market. The matching process in market making can resemble an auction mechanism, where all incoming buy and sell orders are recorded in an order book, sorted by price and then by time. As new orders arrive, they are matched against existing orders in the order book according to price and time priority. If a match is found, the orders are executed, and the order book is updated. The continuous matching of orders facilitates price discovery, reflecting the current supply and demand dynamics. In some markets, periodic auctions may be used to enhance liquidity at specific times (500ms). These auctions aggregate orders over a short period and execute them simultaneously, balancing supply and demand at a single clearing price. This auction-like matching mechanism helps maintain market liquidity and stability, ensuring that trades are executed efficiently and at fair prices.

\textbf{Limited Orderbook.}
In the financial market, the Limit Order Book (LOB) is a crucial component that records all outstanding buy and sell orders. The structure of the LOB can be described using mathematical formulas and symbols to better understand its operational mechanism. The queue position and volume are two important indicators in the LOB, representing the order's position in the queue and the total volume at a specific price level. These are expressed as 
\[
s_q^t = (q_{-K}^t, \ldots, q_{-1}^t, \ldots, q_K^t) \quad s_v^t = (v_{-K}^t, \ldots, v_{-1}^t, \ldots, v_K^t)
\]
where \( q_i^t \) denotes the absolute queue position and \( v_i^t \) denotes the volume at price level \( i \). Additionally, the LOB displays the current market's bid and ask prices along with their respective quantities, represented by matrices:
\[
P_{\text{type}} = (p_{\text{type},1},  \ldots, p_{\text{type},N}) \quad V_{\text{type}} = (v_{\text{type},1}, , \ldots, v_{\text{type},N})
\]
where \(\text{type}\) can be \text{buy} or \text{sell}. The order book depth, indicating the cumulative buy and sell volumes at different price levels, is expressed as 
\[
D_{\text{type}}(p) = \sum_{i=1}^{N} v_{\text{type},i} \quad \text{for} \quad (p_{\text{type},i} \leq p \, \text{and type}) \quad 
\]

\textbf{Order Stacking.}
Order stacking is a trading strategy where market maker submit multiple buy or sell orders in the order book to optimize trade execution. Traders can set multiple orders at different price points, ensuring that these orders are automatically executed when the market price reaches the corresponding levels. This strategy helps capture favorable prices during market fluctuations while managing trading risks. 

Suppose a trader wants to submit ask/bid orders at different price points \( P_1, \ldots, P_n \), with each order having a quantity of \( Q_1, \ldots, Q_n \). The total ask/bid quantity \( Q_{\text{total}} \) can be expressed as:

\[
Q_{\text{total}} = \sum_{i=1}^{n} Q_i
\]

Through this approach, traders can create a price ladder in the order book, allowing for gradual execution of orders as market prices fluctuate, leading to better trading outcomes. The order stacking strategy needs to be dynamically adjusted based on market conditions and trading objectives to maximize returns and control risks.

In the order stacking strategy, order cancellation is performed by traders based on their judgment. When traders decide to cancel orders, they typically do so in a proportional manner. This means that cancellation is carried out proportionally across all pending orders to maintain the consistency of the strategy and the effectiveness of risk management.

Suppose a trader decides to cancel a proportion \( c \) of the total quantity \( Q_{\text{total}} \). The quantity to cancel for each order \( i \) is given by:

\[
Q_{i,\text{cancel}} = c \times Q_i
\]

This approach allows traders to flexibly adjust their orders when market conditions change, ensuring effective risk management and optimization of returns during market fluctuations.

\textbf{Limited Order and Market Order.}
Limited order allows traders to set a specific execution price, and the order will only be executed when the market price reaches or exceeds this price.A major drawback of limit orders is the uncertainty of execution: If the market price does not reach the investor's set limit, the order may not be executed.
Market order is designed to be executed immediately at the current market price, making it particularly suitable for investors who need to quickly enter or exit the market. It typically may cost more than limit orders. This is mainly due to the potential for significant price slippage during execution, especially in highly volatile or illiquid markets. Price slippage refers to the difference between the actual execution price and the expected price.

\textbf{Inventory.}
In market making, inventory refers to the positions held by a market maker after executing trades.
\textbf{$\beta$}
\textbf{Net Value.}
In market making, Net Value refers to the overall value of a market maker's inventory after accounting for any liabilities.

\textbf{Market Making.}
In financial markets, market makers are key participants who facilitate trading and enhance market liquidity by providing continuous ask and bid quotes to the order book. By narrowing the bid-ask spread and supplying liquidity to other market participants, they significantly improve market efficiency. The primary objective of market makers is to maintain a net inventory close to zero, ensuring that the quantity of assets they hold is approximately equal to the quantity sold at any given time. This strategy helps mitigate market risk arising from price fluctuations.

Market makers maintain liquidity by offering two-sided quotes, simultaneously providing a bid price (the price they are willing to pay to buy an asset) and an ask price (the price they are willing to accept to sell an asset). The difference between these prices, known as the bid-ask spread, is the primary source of revenue for market makers. The profit for a market maker can be expressed with the following formula:$
\text{Profit} = \sum_{i=1}^{n} (P_{\text{ask}, i} - P_{\text{bid}, i}) \times Q_i - C
$, where \( P_{\text{ask}, i} \) and \( P_{\text{bid}, i} \) are the ask and bid prices for the \( i \)-th transaction, \( Q_i \) is the quantity of the \( i \)-th transaction, \( C \) represents the costs associated with trading (e.g. fees, taxes) and \( n \) is the total number of transactions.

To effectively control trading prices and risks, market makers often use limit orders to set buy and sell prices. These orders are automatically executed when the market price meets expectations, adhering to the time-price priority in order matching mechanisms. However, when sudden market changes lead to excessive position risk, market makers may need to use market orders to quickly adjust positions. Although market orders can rapidly reduce position risk, they may also increase transaction costs, weakening the market maker's price advantage. Therefore, market makers must balance the use of limit orders and market orders in their trading strategies to optimize liquidity management and risk control.

In addition, the latency between order placement and execution is critical to the effectiveness of market makers. By factoring in latency when determining order elements, they ensure that their quotes remain timely and effective in a rapidly changing trading environment. Market makers dynamically adjust their quotes in response to market fluctuations to attract sufficient trading volume while managing risk exposure. This agility, combined with efficient order matching and execution, is essential for sustaining profitability in a competitive market.

\section{Implementation Details}
\begin{table}[ht]
\centering
\begin{tabular}{c|c}
\toprule
Names & Values \\ \midrule
Learning rate & $1e -3$ \\ 
Entropy coefficient  &  0.001 \\ 
Value function coefficient  & 0.7 \\ 
Number of epochs  & 20 \\ 
Batch size  & 256 \\ 
Discount factor  & 0.99 \\ 
Clip range  & 0.3  \\ 
Max gradient norm   & 0.5  \\ 
Total timesteps  & $1e6$ \\ 
Number of forward steps & 512 \\
\bottomrule
\end{tabular}
\caption{Hyperparameters related to PPO in \textsc{Relaver}}
\label{tab:hyperparameters2}
\end{table}

\begin{table}[H]
\centering
\begin{tabular}{c|c}
\toprule
Names & Values \\ \midrule
Policy Network Hidden Size & 128 \\ 
Policy Network Layers  &  2 \\ 
Value Network Hidden Size & 128 \\ 
Value Network Layers  &  2 \\ 
Optimizer  & Adam \\ 
LSTM Input Size  & 512 \\
LSTM Hidden Size  & 128 \\ 
LSTM Layers & 1 \\
Activation Function & Relu \\
\bottomrule
\end{tabular}
\caption{PPO LSTM Model Structure in \textsc{Relaver}}
\label{tab:hyperparameters}
\end{table}
In \textsc{Relaver}, we implemented the PPO algorithm with an LSTM network structure. Key hyperparameters include a learning rate of $1e-3$, an entropy coefficient of 0.001, and a value function coefficient of 0.7. The training process involved 20 epochs with a batch size of 256 and a discount factor of 0.99. We set the clip range to 0.3 and the maximum gradient norm to 0.5. The total number of timesteps was $1e6$, with 512 forward steps per iteration. For the network architecture, both the policy and value networks consist of 2 hidden layers with 128 neurons each. We used the Adam optimizer and ReLU activation function. The LSTM component has an input size of 512, a hidden size of 128, and a single layer. This configuration balances computational efficiency with the ability to capture temporal dependencies in the MM environment.

\end{document}